\documentclass[conference]{IEEEtran}
\IEEEoverridecommandlockouts
\usepackage{cite}
\usepackage{amsmath,amssymb,amsfonts}
\usepackage{algorithmic}
\usepackage{graphicx}
\usepackage{textcomp}
\usepackage{xcolor}
\usepackage{xspace}
\usepackage{amsmath}
\usepackage{multirow}
\usepackage[ruled]{algorithm2e} 
\usepackage{dblfloatfix}
\usepackage{enumitem}
\def\BibTeX{{\rm B\kern-.05em{\sc i\kern-.025em b}\kern-.08em
    T\kern-.1667em\lower.7ex\hbox{E}\kern-.125emX}}
\begin{document}


\title{

Classifying vaccine sentiment tweets by modelling domain-specific representation and commonsense knowledge into context-aware attentive GRU





\author{\IEEEauthorblockN{Usman Naseem\IEEEauthorrefmark{1},
Matloob Khushi\IEEEauthorrefmark{2}, Jinman Kim\IEEEauthorrefmark{3}, Adam Dunn\IEEEauthorrefmark{4}}

\IEEEauthorblockA{\IEEEauthorrefmark{1}\IEEEauthorrefmark{2}\IEEEauthorrefmark{3}School of Computer Science, The University of Sydney, Australia\\
\IEEEauthorrefmark{4}School of Medical Sciences, The University of Sydney, Australia\\
Email:\{\IEEEauthorrefmark{1}usman.naseem,\IEEEauthorrefmark{2}matloob.khushi,\IEEEauthorrefmark{3}jinman.kim,
\IEEEauthorrefmark{4}adam.dunn\}@sydney.edu.au}}



 
%
}

\maketitle

\begin{abstract}

Vaccines are an important public health measure, but vaccine hesitancy and refusal can create clusters of low vaccine coverage and reduce the effectiveness of vaccination programs. Social media provides an opportunity to estimate emerging risks to vaccine acceptance by including geographical location and detailing vaccine-related concerns. Methods for classifying social media posts, such as vaccine-related tweets, use language models (LMs) trained on general domain text. However, challenges to measuring vaccine sentiment at scale arise from the absence of tonal stress and gestural cues and may not always have additional information about the user, e.g., past tweets or social connections. Another challenge in LMs is the lack of ‘commonsense’ knowledge that are apparent in users’ metadata, i.e., emoticons, positive and negative words etc. In this study, to classify vaccine sentiment tweets with limited information,  we present a novel end-to-end framework consisting of interconnected components that use domain-specific LM trained on vaccine-related tweets and models commonsense knowledge into a bidirectional gated recurrent network (CK-BiGRU) with context-aware attention. We further leverage syntactical, user metadata and sentiment information to capture the sentiment of a tweet. We experimented using two popular vaccine-related Twitter datasets and demonstrate that our proposed approach outperforms state-of-the-art models in identifying pro-vaccine, anti-vaccine and neutral tweets.

\end{abstract}


\section{Introduction}




Vaccines are an effective public health measure used to prevent and control infectious diseases. Access to healthcare services is an ongoing challenge to vaccine coverage, but vaccine hesitancy and refusal can also increase the risk of outbreaks~\cite{19Larson:2010:IIR:1753126.1753129}, and threaten the ability to introduce new vaccines for conditions like human papillomavirus~\cite{zhang2020mining} or COVID-19~\cite{dodd2021concerns,lewandowsky2021covid}. Some of the beliefs or attitudes of people who refuse or are hesitant about vaccination include distrust of government or industry, fear of side effects, lack of effectiveness, or that the disease is not serious~\cite{luz2020heuristics}. 

Social media analysis can identify emerging concerns and the spread of misinformation for use in guiding targeted communication interventions, education and policy~\cite{steffens2020using}. Previous social media research focused on public health topics includes infectious diseases and outbreaks~\cite{charles2015using}, illicit drug use~\cite{kazemi2017systematic}, health promotion~\cite{rao2020mgl,cong2018xa}, and pharmacovigilance support~\cite{golder2015systematic,kang2017semantic}. Twitter is commonly studied because it has a large user base, and data are generally more accessible compared to other platforms~\cite{love2013twitter}. Twitter-specific health research has included analysis of sentiment for health-related issues~\cite{salathe2011assessing,salathe2013dynamics}, including vaccination~\cite{zhang2020sentiment}, allergies~\cite{paul2011you}, and others~\cite{sinnenberg2017twitter,naseem2021covidsenti}.

\begin{figure}[!t]
\centering
\includegraphics[width=1\linewidth]{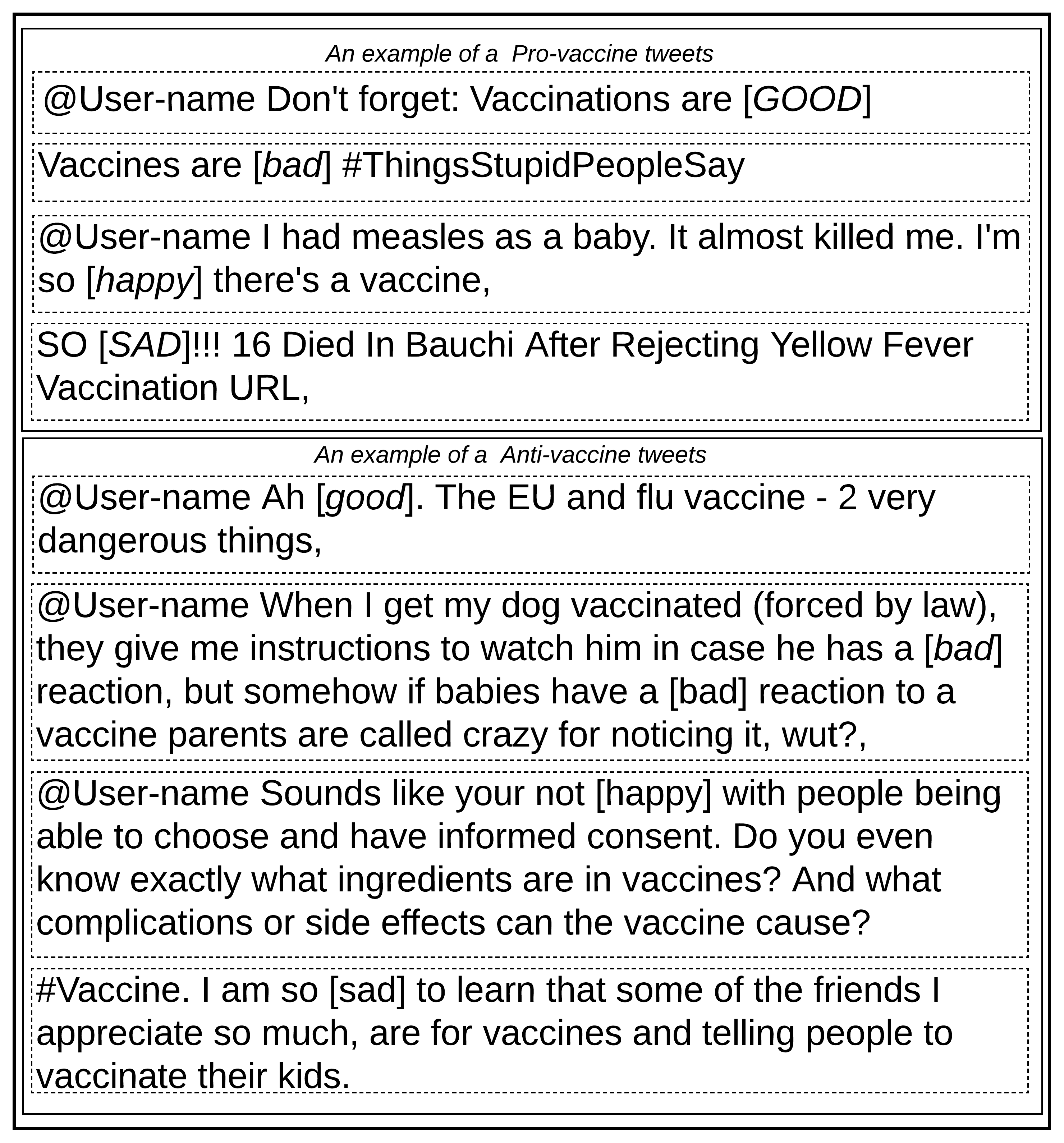}
\caption{ Examples of tweets where meaning of words [good],[bad],[happy], and [sad]  changes according to the context of pro-vaccine or anti-vaccine tweets.}
\label{tweet}
\end{figure}

Twitter has been used to characterize vaccine attitudes among social media users~\cite{dunn2015associations,zhou2015using}.  A broad range of methods has been proposed to classify tweet-level vaccine sentiment based on the user’s expressed vaccine attitudes, often being able to differentiate between vaccine critical tweets or users and neutral or vaccine-promoting tweets or users. Challenges relating to the Twitter format include the short length of the text and informal use of language, abbreviations and misspellings, and inclusion of URLs and emoticons (Fig.~\ref{tweet}).



Deep learning (DL) has played an increasingly important role in natural language processing (NLP)~\cite{naseem2020comprehensive}. Transformer~\cite{vaswani2017attention} based approaches use general language models (LMs) to represent semantic information but can be too general to capture specific context from specific types of text or topics. One limitation of transformer-based models is the inability to capture information that is specific to a domain. To improve results in specialized domains, several transformer-based LMs such as Biomedical BERT (BioBERT)\cite{lee2019biobert}, Biomedical A Lite Bidirectional Encoder Representations from Transformers (BioALBERT)\cite{naseem2020bioalbert}, Twitter BERT (BertTweet)\cite{nguyen2020bertweet} and Covid-BERT (CT-BERT)\cite{muller2020covid}, were trained on domain-specific corpora using the same unsupervised training method used in general models. Given how domain-specific LMs have improved performance for specific NLP tasks, we expect that domain-specific LMs should improve the vaccine sentiment classification task.

Some vaccine sentiment classification methods use information about the social network or users' past tweets to inform the classification of a tweet~\cite{dunn2015associations,zhou2015using}, but this may be infeasible where the number of users posting about vaccines is too large like we see with COVID-19 vaccine tweets currently. A limitation of the research in the area comes from the challenge of incorporating external knowledge (i.e. commonsense knowledge) into the models in ways that could contribute to the performance of the classification task. Commonsense knowledge is described as universally known and acknowledged information about the world~\cite{storks2019recent}.



In this paper, we present an end-to-end method for tweet-level vaccine sentiment classification that models: (i) domain-specific knowledge, (ii) commonsense knowledge, (iii) user metadata, and (iv) word-level sentiment. Our contributions are:

\begin{itemize}[leftmargin=*]
    
    \item We introduce the first large-scale pre-trained language model for English vaccine-related tweets;
    \item We expand the typical Gated Recurrent Unit (GRU)  by introducing a new  cell component that allows for combination with external knowledge and consolidates  commonsense knowledge into our LM network;
    
    \item We present an end-to-end method for vaccine sentiment classification that outperforms recent state-of-the-art (SOTA) methods for tweet-level vaccine sentiment classification.


    
    
    
    
\end{itemize}



\section{Related Work} \label{rw}

Recent advances in NLP have demonstrated a substantial improvement in performance across a range of tasks~\cite{liu2019linguistic,peters2018dissecting}. Context-dependent representations were introduced to address limitations related to capturing context-dependent representations that assign one vector to the same word regardless of context~\cite{Melamud2016context2vecLG,McCann2017LearnedIT,peters2018deep}. Further improvements for specific tasks were achieved by introducing sentiment information extracted from lexicons~\cite{naseem_dice}.

Commonsense knowledge is considered an evident knowledge to humans and contains universally acknowledged opinion about the world~\cite{storks2019recent}. With the advancement of knowledge engineering, there has been a continuous attempt to collect and encode commonsense knowledge, and many commonsense knowledge bases (CKBs) such as ConceptNet, FreeBase, DBpedia, and SenticNet have been used to augment models with real-world knowledge for different NLP tasks. For example, Ahn et al.~\cite{ahn2016neural} developed a LM that leveraged knowledge bases in an RNN-based LM. Ye et al.~\cite{ye2019align} proposed a discriminative pre-training method for including commonsense knowledge into
the LM, in which the question is concatenated
with different candidates to build a multi-choice question answering sample, and each choice is used to predict whether the candidate is the right answer. For sentiment analysis, Xu et al.~\cite{xu2017incorporating} modified the ordinary recall gate function in RNN to leverage CKB. For sentiment classification, Ma et al.~\cite{ma2018targeted} combined  CKB into LSTM cell to improve aspect sentiment classification. Unlike their work, we leverage the commonsense knowledge in BiGRU with context-aware attention and applies it to information extraction.

Most tweet-level vaccine sentiment classification methods use traditional machine learning methods include support vector machines (SVM)~\cite{botsis2011text}, Naive Bayes ensembles and maximum entropy classifiers~\cite{salathe2013dynamics}, and hierarchical SVMs~\cite{du2017optimization}. Alternative approaches use the information beyond individual tweets, including social network structure~\cite{dunn2015associations,zhou2015using}. As supervised machine learning methods, these approaches rely on manual labour from experts. Zhang et al.~\cite{zhang2020sentiment} used three transfer learning methods (ELMo, GPT, and BERT) for tweet-level HPV vaccine sentiment classification and found that a fine-tuned BERT model produced the highest performance.

\begin{figure}[!t]
\centering
\includegraphics[width=.98\linewidth]{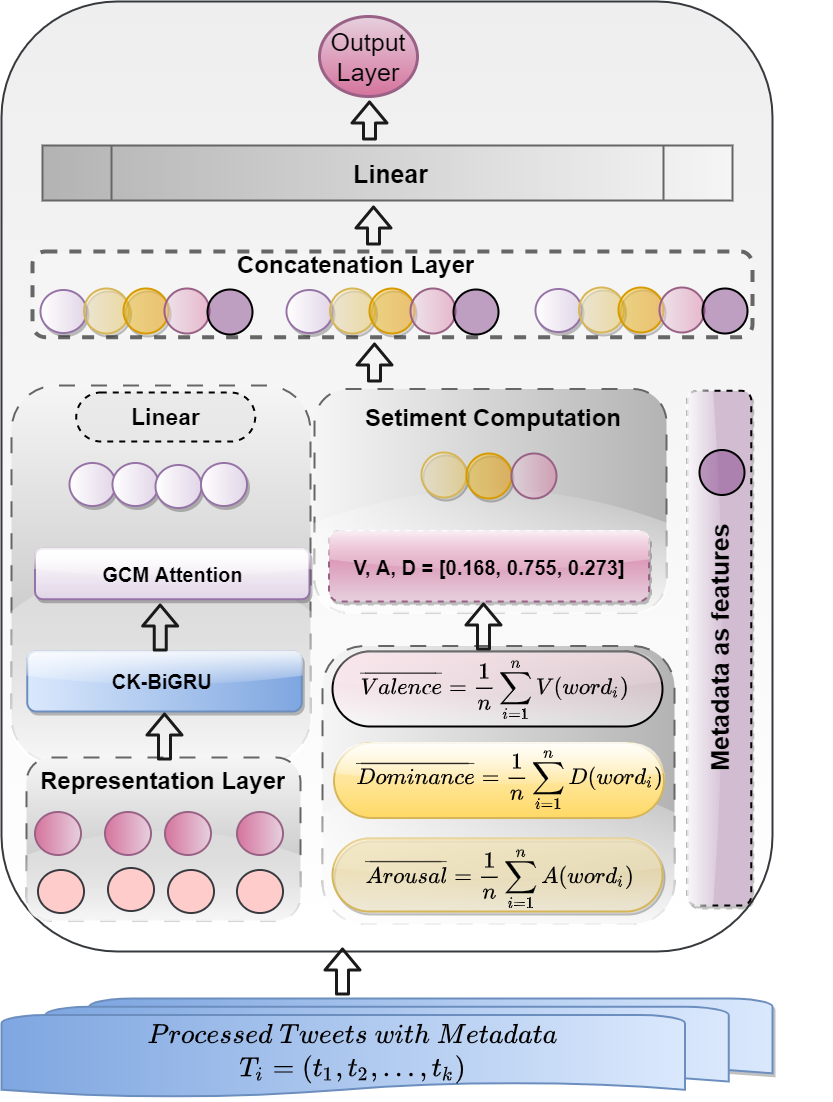}
\caption{ Overall architecture of the proposed approach to tweet-level vaccine sentiment classification.}
\label{arciiii}
\end{figure}

\section{Methodology}\label{111model}

First, we define our target problem, followed by the architecture of the proposed method and technical details.


\textbf{Problem Definition}: First we define our problem formally, given a \textbf{tweet} \( T_i \) with a sequence of tokens \( (t_1,t_2,t_3,...t_k ) \), \textit{i}  describes the number of a \textbf{tweet} and \textit{k} shows the number of tokens in a \textbf{tweet}. The objective is to predict whether a tweet sentiment polarity is positive, negative or neutral from corresponding set of
labels \(Y = (y_1,y_2,...,y_k)\).

\textbf{Overview of Architecture}:
The model is based on (a)
Representation layer, which includes domain-specific contextual word representation and linguistic-based embedding; (b) Bidirectional Gated Recurrent Unit with Commonsense (CK-BiGRU); (c) Context-aware attention; (d) Sentiment Embedding; and (e) User behaviour features. The architecture of the proposed model can be seen in Fig.~\ref{arciiii}. In the subsequent discussion, we will explain each of these components in depth.

\subsection{Representation Layer}

As mentioned previously, the first component of our model is a representation layer that includes word representations obtained using domain-specific pre-trained LM and linguistic embedding. Below we describe each of these.

\textbf{RoBERTa for Vaccine related tweets:} A Robustly Optimized BERT (RoBERTa)~\cite{robertaliu2019roberta} is an optimized version of Bidirectional encoder representations from transformer (BERT)~\cite{devlin-etal-2019-bert}. During pre-training, BERT used two training objectives: (i) mask language modelling (MLM) and (ii) a next sentence prediction (NSP) task, whereas RoBERTa made the following changes to the BERT model: (i) Longer training of the model with more data with bigger batches; (ii) Eliminating the NSP objective; (iii) Longer training of the sequence; (iv) Dynamically changing the masked positions during pre-training. 
Our sub-domain contextual LM uses the same architecture as RoBERTa and is trained on a corpus of 64M vaccine-related tweets crawled using Twitter API during January 12, 2017, and December 3, 2019, using keywords related to the vaccine in the English language.

We used the same pre-processing techniques before training as used in previous studies~\cite{muller2020covid,nguyen2020bertweet}. The retweet tags were removed from the raw corpus, and usernames and web-page URLs were replaced with unique tokens @USER and HTTP-URL, respectively. Further, all emoticons were replaced with their associated meaning using the Python emoji library\footnote{https://pypi.org/project/emoji/}. Tweets were segmented using open-source python, the HuggingFace library\footnote{https://huggingface.co/}.  Every input sequence of the RoBERTa LM is transformed into 50,265-word vocabulary tokens. The length of  Twitter messages is limited to 200 characters, and we kept the batch size of 8 during the training and evaluation process.

\textbf{Linguistic Embedding:} Each word in a sentence has several attributes which we can use for analysis. For instance, part of speech (POS) of a token where nouns are a person, location, or object; verbs are acts or events; adjectives are terms that characterise nouns. However, words often have relations around them, and there are a variety of such relations. For instance, a noun may be the subject of a sentence where it behaves as a verb, e.g.\textit{`Trump laughed'}. Nouns may also be the subject of a sentence where they are the subject of a sentence, such as \textit{Biden} in the example \textit{`Trump laughed at Biden'}. Word representation as discrete and distinct symbols is incomplete and often leads to poor generalisation. Thus, we
aimed to produce a representation that preserves semantic and syntactive similarity between words using linguistic embedding.

Here, we used both \textit{POS} \textit{tags} to provide knowledge of a word and the various POS forms of words and \textit{dependency parsing} to explain these relations among words in a phrase to model linguistic features in a tweet context.  Using POS tagging and dependency parsing has demonstrated positive results in previous studies~\cite{naseem2020transformer}. We used the Python library `SpaCy'\footnote{https://pypi.org/project/spacy/} for dependency and POS tagging. These tags were then transformed into a vector representation using one-hot encoding to capture syntactical information of words. 




\subsection{BiGRU with commonsense knowledge (CK-BiGRU)}
\textbf{GRU:}  is a type of RNNs presented by  \cite{184DBLP:journals/corr/ChungGCB14}. GRU is the simplest form of LSTM architecture. It includes two gates and does not contain internal memory, which makes it different from LSTM. Also, in GRU, a second non-linearity (tanh) is not applied on a network. The working of a GRU cell is given below: 
    

    
    \(z_{t}= \sigma(W_{iz}x_{t}+b_{iz}+W_{hz}h_{(t-1)}+b_{hz})\)


    
    \( r_{t} = \sigma(W_{ir}x_{t}+b_{ir}+ W_{hr}h_{(t-1)}+b_{hr})\)
    

    
   \(n_{t} = \tanh (W_{in}x_{t}+b_{in}+t_{t}*W_{hn}h_{(t-1)}+b_{hn})\)
   
    \(h_{t} = (1- z_{t})*n_{t}+z_{t}*h_{(t-1)}\) 
   
   Where  \(z_{t}\) is the reset, \(r_{t}\) is the update, \(n_{t}\) is the new gates, $\sigma$ is the sigmoid function, and * is the product of Hadamard, and \(x_{t}\) and \(h_{t}\) is the input and the hidden state at time \textit{t}, and finally, \(h_{(t-1)}\) is the hidden state of the layer at time \textit{t-1}. Typically, a single GRU encodes a sequence in only one direction.However, two GRUs can be stacked to use it as a bi-directional encoder, attributed to as a bi-directional GRU that leads to two hidden states forward \(\overrightarrow{h_i}\) and backward \(\overleftarrow{h_i}\) GRU at time step \textit{t}. Their concatemnation  $h_i= [ \overrightarrow{h_i} \parallel \overleftarrow{h_i}] $ provides a full description of the input data for the time step \textit{t}.
   
   

\begin{figure}[!t]
\centering
\includegraphics[width=.98\linewidth]{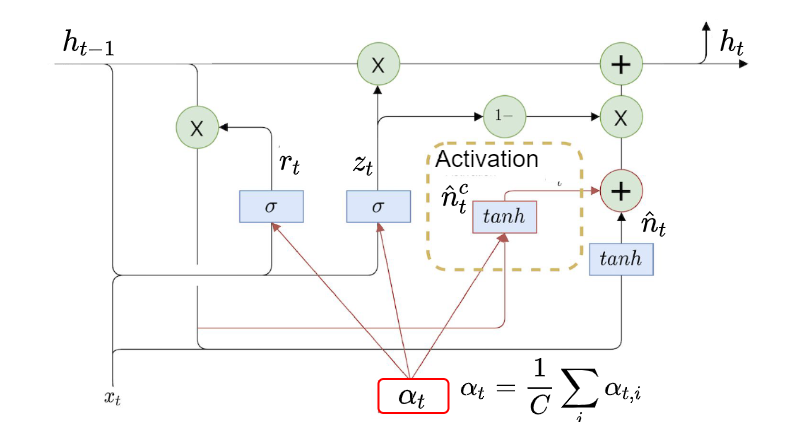}
\caption{A Visualisation of BiGRU with commonsense knowledge (CK-BiGRU).}
\label{comm}
\end{figure}

   
\textbf{Commonsense Knowledge:} To improve performance, we used commonsense knowledge as our knowledge base to be incorporated in the sequence encoder.  Typically, a CKB can be considered as a semantic network where concepts are nodes in the graph and relations are edges. Each~ \(<~concept1,~relation,~concept2~>\) triple is termed an assertion.  AffectiveSpace~\cite{sentic5.Cambria2018SenticNet5D} was designed to map SenticNet~\cite{sentic5.Cambria2018SenticNet5D} CKB concepts to continuous low-dimensional embeddings while maintaining the affective relationship of the original space. We used AffectiveSpace and extend a previous approach~\cite{ma2018targeted} to construct an affective extension of BiGRU called BiGRU with commonsense knowledge (CK-BiGRU). CK-BiGRU leverages the affective commonsense knowledge to GRU cells and provides affective information to the GRU memory cell. A set of \textit{C} concept candidates would be extracted using a
syntactic concept parser and mapped to the $d_{c}$ dimensional vectors \( [\alpha_{t,1}, \alpha_{t,2}, ....., \alpha_{t,C}]
\) at time step \textit{t}. The candidate embedding of step \textit{t} is calculated as the average of the vectors using  eq\ref{cc}:
 \begin{equation}\label{cc}
      \alpha_{t}= \frac {1}{ C}    \sum_{i} \alpha_{t,i}
 \end{equation}

The formula of BiGRU with commonsense knowledge is illustrated below :

    \( r_{t} = \sigma(W_{r}[x_{t}, h_{t-1},\textcolor{red}{\alpha_{t}}]+b_{r})\)
    
     \( z_{t} = \sigma(W_{z}[x_{t}, h_{t-1},\textcolor{red}{\alpha_{t}}]+b_{z})\)
    

     \(\widehat{n_{t}} = \tanh (W_{n}x_{t}+b_{n}+r_{t}*(W_{m}x_{t}+b_{m}))\)
     
     \(\textcolor{red}{\widehat{n_{t}^{c}} = \tanh (W_{cn}\alpha_{t}+b_{cn}+r_{t}*(W_{cm}x_{t}+b_{cm}))}\)

   
    \(h_{t} = (1- z_{t})*n_{t}+z_{t}*h_{(t-1)}+\textcolor{red}{(1-z_{t})*\widehat{n_{t}^{c}}}\) 


Knowledge concepts are added to the reset and update gates as a filtering cue. They are
assumed to be meaningful to the sequence model to supervise the information flow at the token level. Moreover, an additional candidate activation vector \(\widehat{n_{t}^{c}}\) models the relative contributions of token and concept level, is employed to extend the normal GRU cell and is added to the output vector, as shown in Fig~\ref{comm}.



\subsection{Context-Aware Attention}

Not all words have an equivalent role in interpreting the context of a sentence, and attention  assigns a weight \textit{\(a_n\)} to each word through a softmax function. In this work, we used global context-aware attention   mechanism (GCM)~\cite{liu2017global} to impose the contribution of meaningful words instead of simply
using CK-BiGRU hidden states, since the outputs of the state provide relatively local contextual information
from previous time steps.



\begin{figure}[!t]
\centering
\includegraphics[width=.98\linewidth]{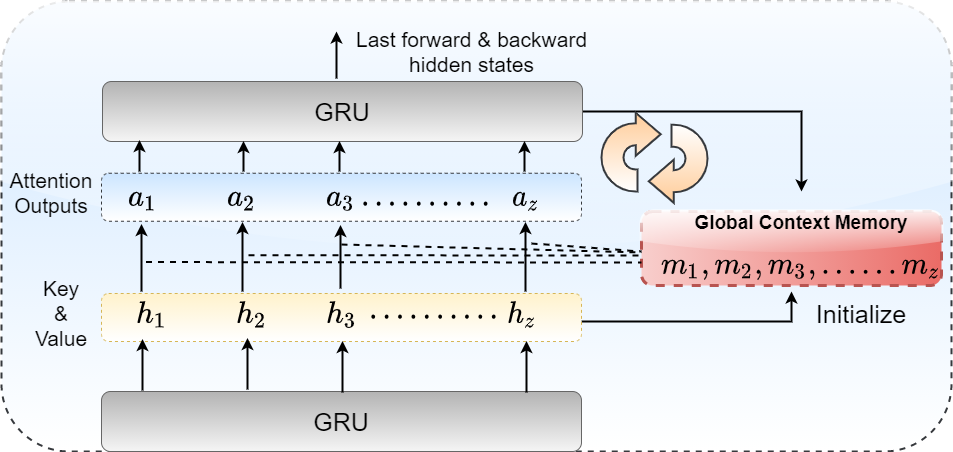}
\caption{ A Visualisation of Global Context-aware Attention.}
\label{gcm}
\end{figure}

The hidden states of all time steps \( [h_{1},h_{2},.....,h_{n}]\) from the first CK-BiGRU layer act as the key and value vectors, while the query vectors are from the GCM. The vectors list \( [h_{1},h_{2},.....,h_{n}]\)   also initializes the memory \([m_{1}, m_{2},.....,m_{n}] \). The
attention outputs \([a_{1},a_{2},....,a_{n}] \) are fed into the second CK-BiGRU layer and then the state outputs refine the GCM. Multiple iterations of attention operations would be conducted to promote the attention ability to assess each token's informativeness level. Liu et al.\cite{liu2017global} use the memory for the final classification. However, in our model, the last forward and backward hidden outputs from the second CK-BiGRU layer are generated as the entire attention layer's output after iterative refinements. A visual representation of GCM attention is shown in Fig.~\ref{gcm}. We fed the representation of CK-BiGRU and context-aware attention into a linear layer. The linear layer aims to reduce the dimensionality, to avoid weakening the influence of small-dimensional sentiment scores on the final results.

\subsection{Sentiment Lexicon}

Extending our method, we incorporated a sentiment score for the tweet into our model.  The sentiment embedding obtains sentiment scores from lexicons. Each lexicon includes a pair of words and their corresponding sentiment, in which each word has its sentiment value.


We used Valence Arousal Dominance (VAD) sentiment lexicon~\cite{mohammad2018obtaining}  to extract the sentiment scores and is transformed into a vector that captures words' sentiment.   The VAD lexicon gives a vector,  \(\overrightarrow {VAD_{t}} = [t_{V}, t_{A}, t_{D}]\), for every token where \(t_{V}\) is the valence of the token,  \(t_{A}\) is the arousal of the token, and \(t_{D}\) is the dominance of the token. These three concepts have been recognized as the most significant aspects of meaning by a number of linguistic scholars ~\cite{mohammad2018obtaining}. Our VAD-lexicon-based sentiment calculates each dimension's average score (valence, arousal and dominance) for each part separately and returns the three scores corresponding to each part. A value of zero is given to a word that is not present in the lexicon. The output would be a 9-dimensional vector containing 3 score lists for each part of a tweet.

\subsection{Incorporating User-Metadata}

Many new social media sentiment analysis approaches evaluate 
tweet-level vaccine sentiment mainly on the basis of textual content and neglect other useful information (metadata such as click, follow, \# of friends and followers, etc.) easily available on these platforms. Incorporating metadata has demonstrated positive results in previous studies~\cite{alharbi2019twitter}. In our work, we tested a combination of different metadata (given in Table~\ref{metaa})  as features to train our model. The feature is processed for the entire list with min-max normalisation since different tweets contain quite different amounts of those particles. The best combination (F1-F8) of metadata is obtained from testing different combinations (see Table IV).




\begin{table}[!t]
\centering
\caption{List of features used to incorporate User's behaviour in metadata}
\label{metaa}
\begin{tabular}{cl}
\hline
Feature ID & \multicolumn{1}{c}{Feature Description}                        \\ \hline
F1         & Tweet posted Date,                                             \\ \hline
F2         & \# of Emoticons,                                           \\ \hline
F3         & \# of Hashtags,                                            \\ \hline
F4         & \# of Exclamation Marks,                                   \\ \hline
F5         & \# of Question Marks,                                      \\ \hline
F6         & \# of Mentions,                                            \\ \hline
F7         & \# of Positive Words in Bing Liu Lexicon, \\ \hline
F8         & \# of Negative Words in Bing Liu Lexicon, \\ \hline
F9         & \# of Favorite                                             \\ \hline
F10        & \# of Retweet                                              \\ \hline
F11        & \# of User’s Favorite                                      \\ \hline
F12        & \# of Followers                                            \\ \hline
F13        & \# of Friends                                              \\ \hline
F14        & \# of User Listed                                          \\ \hline
F15        & User Statues                                                   \\ \hline
F16        & Profile Verified (Yes/No)                                                           \\ \hline
F17        & Profile Image (Yes/No)                                                 \\ \hline
\end{tabular}
\end{table}

Finally, we combined the above vectors to produce a vector that can resolve the language ambiguities described above. The final concatenated vector is then fed to the linear output layer for final prediction. We used softmax to have the distribution of the likelihood class at the last layer of the classifier and used cross-entropy as a loss function in our experiments.

\begin{table}[!b]
\centering
\caption{Dataset Distribution}
\label{data11}
\begin{tabular}{ccc}
\hline
Dataset               & Vaccine Sentiment (VS1) & Vaccine Sentiment (VS2) \\ \hline
Positive & 6,683                   & 8,965                   \\ \hline
Negative       & 1,084                   & 1,976                   \\ \hline
Neutral               & 1,445                   & 7,562                   \\ \hline
Total                 & 9,212                   & 18,503                  \\ \hline
\end{tabular}
\end{table}

\section{Experiments}\label{results}

We evaluated the performance of our approach with several SOTA approaches as the baselines. A 10-fold cross-validation technique is used to evaluate the classification results. Accuracy, F1-Score, Recall and Precision scores are reported to evaluate the performance of our model.





\begin{table*}[!t]
\centering
\caption{Comparison of the proposed method v/s the baselines, F1-score, Precision and Recall score averaged over 10-folds.}
\label{results}
\begin{tabular}{ccccc|cccc}
\hline
\multirow{4}{*}{Model\textbackslash Dataset} & \multicolumn{4}{c|}{\multirow{2}{*}{VS1}}                                                                    & \multicolumn{4}{c}{\multirow{2}{*}{VS2}}                                                                     \\
                                                            & \multicolumn{4}{c|}{}                                                                                        & \multicolumn{4}{c}{}                                                                                         \\ \cline{2-9} 
                                                            & \multirow{2}{*}{Accuracy} & \multirow{2}{*}{F1-Score} & \multirow{2}{*}{Precision} & \multirow{2}{*}{Recall} & \multirow{2}{*}{Accuracy} & \multirow{2}{*}{F1-Score} & \multirow{2}{*}{Precision} & \multirow{2}{*}{Recall} \\
                                                            &                           &                           &                            &                         &                           &                           &                            &                         \\ \hline
Word2vec                                                    & 0.718                     & 0.624                     & 0.611                      & 0.718                   & 0.639                     & 0.610                     & 0.615                      & 0.639                   \\ \hline
Glove                                                       & 0.730                     & 0.652                     & 0.672                      & 0.720                   & 0.676                     & 0.643                     & 0.674                      & 0.696                   \\ \hline
BERT                                                        & 0.746                     & 0.682                     & 0.706                      & 0.746                   & 0.751                     & 0.738                     & 0.746                      & 0.751                   \\ \hline
RoBERTa                                                     & 0.764                     & 0.714                     & 0.733                      & 0.764                   & 0.765                     & 0.757                     & 0.762                      & 0.765                   \\ \hline
ALBERT                                                      & 0.747                     & 0.681                     & 0.711                      & 0.747                   & 0.739                     & 0.726                     & 0.732                      & 0.739                   \\ \hline
BioBERT                                                     & 0.738                     & 0.664                     & 0.690                      & 0.738                   & 0.741                     & 0.725                     & 0.737                      & 0.741                   \\ \hline
BERTweet                                                    & 0.749                     & 0.681                     & 0.719                      & 0.749                   & 0.750                     & 0.737                     & 0.747                      & 0.750                   \\ \hline
CT-BERT                                                     & 0.789                     & 0.752                     & 0.760                      & 0.789                   & 0.783                     & 0.779                     & 0.782                      & 0.783                   \\ \hline
CT-BERT+ Dep   Emb+ BiGRU +   Context Attention             & 0.793                     & 0.778                     & 0.771                      & 0.793                   & 0.797                     & 0.796                     & 0.798                      & 0.797                   \\ \hline
Word2vec+ Dep   Emb+ BiGRU +   Context Attention +VAD       & 0.718                     & 0.683                     & 0.670                      & 0.718                   & 0.731                     & 0.729                     & 0.730                      & 0.731                   \\ \hline
Glove+   Dep Emb+ BiGRU +   Context Attention +VAD          & 0.731                     & 0.706                     & 0.697                      & 0.731                   & 0.741                     & 0.740                     & 0.742                      & 0.731                   \\ \hline
BERT+ Dep   Emb+ BiGRU + Context   Attention +VAD           & 0.750                     & 0.729                     & 0.720                      & 0.750                   & 0.764                     & 0.764                     & 0.764                      & 0.764                   \\ \hline
RoBERTa+   Dep Emb+ BiGRU +   Context Attention +VAD        & 0.760                     & 0.747                     & 0.739                      & 0.760                   & 0.782                     & 0.782                     & 0.784                      & 0.782                   \\ \hline
ALBERT+ Dep   Emb+ BiGRU +   Context Attention +VAD         & 0.745                     & 0.724                     & 0.715                      & 0.745                   & 0.764                     & 0.764                     & 0.764                      & 0.764                   \\ \hline
BioBERT+   Dep Emb+ BiGRU +   Context Attention +VAD        & 0.734                     & 0.718                     & 0.710                      & 0.734                   & 0.754                     & 0.754                     & 0.757                      & 0.754                   \\ \hline
BERTweet+ Dep   Emb+ BiGRU +   Context Attention +VAD       & 0.761                     & 0.742                     & 0.734                      & 0.761                   & 0.773                     & 0.774                     & 0.775                      & 0.773                   \\ \hline
CT-BERT+ Dep   Emb+ BiGRU +   Context Attention +VAD        & 0.792                     & 0.778                     & 0.771                      & 0.792                   & 0.797                     & 0.796                     & 0.798                      & 0.797                   \\ \hline \hline
Proposed                                                    & \textbf{0.820 }                    & \textbf{0.807 }                    & \textbf{0.801 }                     & \textbf{0.820 }                  & \textbf{0.818 }                    & \textbf{0.819 }                    & \textbf{0.823}                      & \textbf{0.818 }                  \\ \hline \hline
\end{tabular}
\end{table*}

\subsection{Datasets}

The experiments were performed using 2 datasets of vaccine-related tweets, where class labels were positive, neutral, and negative (see Table ~\ref{data11}).


\begin{itemize}
    
    \item \textbf{Vaccine Sentiment \#1 (VS1)}: Our first dataset contains the collection of tweets related to the dissemination of vaccine on Twitter by assessing awareness and interaction among regular users in the United States (US) collected between January 12, 2017, and December 3, 2019. This dataset was collected and labelled by~\cite{dunn2020limited}. The total number of tweets are 9,212 and contains three classes: promoting vaccination (positive with 6,683 tweets), vaccine critical (negative with 1,084 tweets) and neural with 1,445 tweets.

    

\item \textbf{Vaccine Sentiment \#2 (VS2)}: This dataset \footnote {https://github.com/digitalepidemiologylab/crowdbreaks-paper} contains a set of measles and vaccination-related U.S.-geolocated tweets gathered between July 2018 and January 2019 via the Twitter Streaming API, presented by Müller et al.~\cite{muller2019crowdbreaks} and contains three classes. The total number of tweets are 18,503 and contains three classes: pro-vaccine (positive with 8,965 tweets), anti-vaccine (negative with 1,976 tweets) and neural with 7,562 tweets. 
\end{itemize}



\subsection{Experimental Settings}

\subsubsection{Pre-processing}
We corrected for spelling mistakes; sentiment aware tokenization was used to replace emojis with their associated words; further, we replaced emoticons with their associated meaning words following previous studies~\cite{naseem2020survey}. With hashtags, the hashtag symbol (\#) was removed, and we achieved word-segmentation to split the words in the hashtag. Other corrections included expanding contractions, normalized URLs, digits, emails, user mentions and elongated words. We used \textit{ekphrasis}\footnote{https://github.com/cbaziotis/ekphrasis} and \textit{emoji}\footnote{https://pypi.org/project/emoji/}, a Python open-source library to enhance the quality of tweets.  We also removed punctuation and removed repeated words using a regular expression.

\begin{table*}[!b]
\centering
\caption{Comparison of our approach with other variants by replacing CK-BiGRU with BiGRU and Sentic-LSTM and different combination of metadata features (F1-F17) given in Table\ref{metaa}.}
\label{Variants}
\begin{tabular}{ccccc|cccc}
\hline
\multirow{2}{*}{Model/Dataset}       & \multicolumn{4}{c|}{VS1}                 & \multicolumn{4}{c}{VS2}                  \\ \cline{2-9} 
                                     & Accuracy & F1-Score & Precision & Recall & Accuracy & F1-Score & Precision & Recall \\ \hline \hline
Proposed Model (PM)                                   & \textbf{0.820}    & \textbf{0.807}    & \textbf{0.801 }    & \textbf{0.820 } & \textbf{0.818 }   & \textbf{0.819}    & \textbf{0.823 }    & \textbf{0.818}  \\ \hline \hline
PM-metadata-CKBiGRU+BiGRU            & 0.788    & 0.779    & 0.773     & 0.788  & 0.795    & 0.796    & 0.799     & 0.795  \\ \hline
PM-CKBiGRU+BiGRU+Metadata (F1-F8)    & 0.798    & 0.783    & 0.776     & 0.798  & 0.797    & 0.796    & 0.797     & 0.797  \\ \hline
PM-CKBiGRU+BiGRU+Metadata (F2-F8)    & 0.786    & 0.780    & 0.775     & 0.786  & 0.796    & 0.797    & 0.800     & 0.796  \\ \hline
PM-CKBiGRU+BiGRU+Metadata (F1-F6)    & 0.790    & 0.778    & 0.772     & 0.790  & 0.791    & 0.792    & 0.795     & 0.791  \\ \hline
PM-CKBiGRU+BiGRU+Metadata (F7-F8)    & 0.792    & 0.778    & 0.771     & 0.792  & 0.797    & 0.796    & 0.798     & 0.797  \\ \hline
PM-CKBiGRU+BiGRU+Metadata (F1-F6)    & 0.792    & 0.778    & 0.771     & 0.792  & 0.796    & 0.796    & 0.797     & 0.796  \\ \hline
PM-CKBiGRU+BiGRU+Metadata (F1-F17)   & 0.790    & 0.780    & 0.774     & 0.790  & 0.796    & 0.796    & 0.798     & 0.796  \\ \hline
PM-CKGRU+SenticLSTM+Metadata (F1-F8) & 0.795    & 0.780    & 0.778     & 0.795  & 0.799    & 0.798    & 0.801     & 0.799  \\ \hline
\end{tabular}
\end{table*}

\subsubsection{Parameters tuning}


We used the grid search optimization method to tune the optimal model. The $(0.5)$ dropout at network connections is used to switch off the network neurons randomly. To eliminate over-fitting and make our approach robust, we used the $(L {2}=0.005)$ regularisation method to determine large weights. We used 2 layers, and the size of each hidden layer used is 50. The classifier was trained with 128 batch size, and to tune the $(0.005)$ learning rate, we used the Adam Optimizer for 40 epochs. Finally, in our method, we used the cross-entropy loss function.

\subsection{Baselines}

We compared our results with several word representation methods and SOTA architectures that have been excessively used for Twitter sentiment analysis. We also evaluate the results with a number of variants of our proposed approach to highlight our technical contribution. We used our implementation of these baselines and used the grid-search cross-validation to derive the models' optimal settings.

\begin{table*}[!t]
\centering
\caption{Comparison of Prediction by our Proposed method in contrast to gold truth.}
\label{predictions}
\begin{tabular}{lcc}
\hline
\multicolumn{1}{c}{Tweet}                                                                 & Gold Truth   & Predicted Label \\ \hline \hline
The same people who mock anti   vaxxers are the same ones who want open borders. \#elpaso & Neutral      & Neutral         \\
$ @username$ Getting my puppers   vaccinated. Safe for another year.                    & Positive  & Positive     \\
Peep dont realize how   dangerous these vaccinations are. https://t.co/VY2ht6ub5Y         & Negative & Negative    \\
The vaccines today are so   dangerous and obviously worthless!! https://t.co/x8CZuX4c8n   & Negative & Negative    \\ \hline \hline
\end{tabular}
\end{table*}

We used representations from pre-trained  Word2Vec~\cite{NIPS2013_5021}, GloVe~\cite{pennington2014glove}, BERT~\cite{devlin-etal-2019-bert}, RoBERTa~\cite{robertaliu2019roberta}, a distilled version of BERT (DistilBERT)\cite{DBLP:journals/corr/abs-1910-01108} and    A Lite Bidirectional Encoder Representations from Transformers (ALBERT)~\cite{lan2019albert}, BioBERT~\cite{lee2019biobert}, BERTTweet~\cite{nguyen2020bertweet} and CT-BERT~\cite{muller2020covid} and fed these representations to a linear classifier. In addition, we also compared our results with neural network-based classifiers with attention. We used the representations from the same word representation models and linguistic features with BiGRU with context-aware attention with and without sentiment embedding. Finally,  we compared the results (Table V) of our approach with other variants by replacing different components such as CK-BiGRU with BiGRU and Sentic LSTM~\cite{ma2018targeted} and different combination of metadata.

\subsection{Results and Discussion}

The proposed method achieved an F1-score of 0.807 on the VS1 dataset and 0.819 on the VS2 dataset, compared to the next highest performance of 0.778 and 0.798, respectively (Table~\ref{results}).

Replacing CK-BiGRU with BiGRU reduced the performance from 0.807 to 0.779 on the VS1 dataset and from 0.819 to 0.795 on the VS2 dataset, showing that CKBiGRU contributed to the overall performance (Table~\ref{Variants}). Similarly, replacing CKBiGRU with SenticLSTM reduced the performance from 0.807 to 0.780 on the VS1 dataset and from 0.819 to 0.798 on the VS2 dataset (Table~\ref{Variants}). Finally, replacing the combination of (F1-F8) metadata features with other combination of features also reduced the performance, showing that the combination of (F1-F8) metadata features also contributed to the overall performance (Table~\ref{Variants}). Overall, the results show that the use of CKBiGRU, a combination of (F1-F8) metadata, and domain-specific contextual word representation together lead to an increase in performance.

\begin{figure}[!htpb]
\centering
\includegraphics[width=1\linewidth]{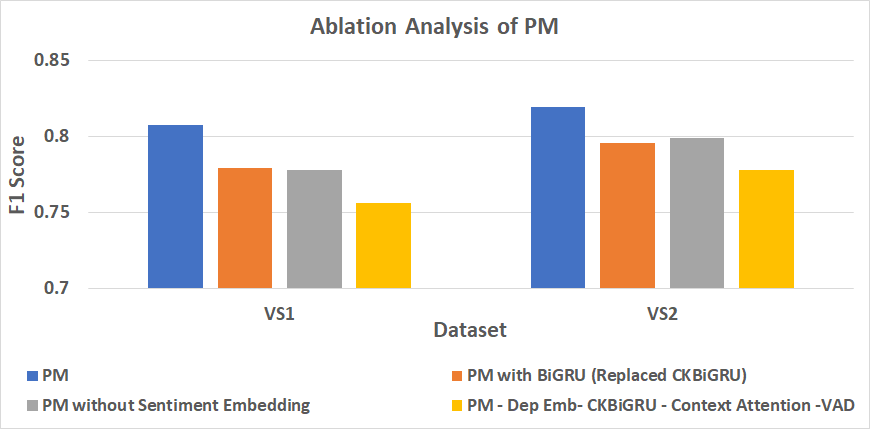}
\caption{An ablation analysis}
\label{ablation}
\end{figure}

\subsection{Ablation Analysis}

Similarly, the ablation testing showed that it is the combination of components that produces the increase in performance (Fig.~\ref{ablation}). The results drop marginally in both cases when we exclude sentiment and replace CK-BiGRU with BiGRU from our approach. Further, the empirical analysis also reveals a noticeable drop in performance when we use our pre-trained domain-specific LM as a linear classifier and remove all other components from our model.



\begin{figure}[!hpbt]
\centering
\includegraphics[width=1\linewidth]{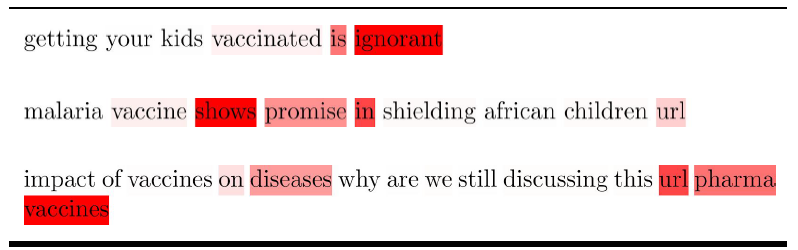}
\caption{Visualisation of GCM Attention heat-map: The color intensity corresponds to the weight given to each word by the GCM-attention mechanism.}
\label{attheatmap}
\end{figure}


An attention-based heat-map visualization illustrated the specific words that are more important to the prediction (Fig.~\ref{attheatmap}). Examples of the predictions made by the proposed approach also illustrate the performance (Table~~\ref{predictions}).







\section{Conclusion}\label{con}

In this study, we presented an end-to-end approach to tweet-level vaccine sentiment classification and demonstrated that our approach outperformed comparative SOTA approaches. Ablation testing shows that each of the components in the proposed approach contributes to the increase in performance over previous approaches. We explicitly modelled the domain-specific contextual word representation with commonsense knowledge and sentiment information, generating a more accurate representation. A key contribution of the work is the first large-scale pre-trained LM for English language vaccine-related tweets. In addition, the novel CK-BiGRU extension demonstrates how commonsense knowledge can be incorporated into a vector. The implications of this work include ways to improve public health surveillance related to vaccine hesitancy and may be extended to support other applications where public opinion has the potential to affect population health outcomes. 

\section{Acknowledgment}
This research is supported by  Australian Government Research Training Program (RTP).

\bibliographystyle{plain} \bibliography{bibo.bib} 

\end{document}